\pdfoutput=1
\documentclass[11pt]{article}

\usepackage[preprint]{acl}
\usepackage{times}
\usepackage{latexsym}
\usepackage[T1]{fontenc}
\usepackage[utf8]{inputenc}

\usepackage{microtype}
\usepackage{tikz}
\usepackage{subcaption}
\usepackage{booktabs} %
\usepackage{multirow}
\usepackage[most]{tcolorbox}

\usepackage[utf8]{inputenc}

\usepackage{xcolor}

\newtcblisting{promptbox}{
  colback=black!3,
  colframe=black!30,
  boxrule=0.4pt,
  arc=2pt,
  left=6pt,right=6pt,top=4pt,bottom=4pt,
  listing only,
  listing options={
    basicstyle=\ttfamily\small,
    breaklines=true,
    columns=fullflexible
  },
  breakable,
  enhanced,
  pad at break=2mm,
}

\usepackage{ragged2e}
\usepackage{tabularx}
\newcolumntype{Y}{>{\RaggedRight\arraybackslash}X}

\definecolor{myblue}{RGB}{70, 130, 180}
\definecolor{myred}{RGB}{220, 60, 60}
\definecolor{mygreen}{RGB}{60, 160, 60}
\definecolor{myorange}{RGB}{230, 140, 0}
\definecolor{bglight}{RGB}{245, 247, 250}

\usepackage{amsmath}
\usepackage{amssymb}
\usepackage{mathtools}
\usepackage{amsthm}

\usepackage[capitalize,noabbrev]{cleveref}

\theoremstyle{plain}

\theoremstyle{definition}

\theoremstyle{remark}

\usepackage[disable,textsize=tiny]{todonotes}

\title{Joint Constraint Aggregation for \\Qualifier-Sensitive Retrieval-Augmented QA}

\author{
Jennifer Hsia \and Zachary Chase Lipton \and Aarti Singh \\
Machine Learning Department \\
Carnegie Mellon University \\
Pittsburgh, PA 15213, USA \\
\texttt{jhsia2@cs.cmu.edu} \quad
\texttt{zlipton@cmu.edu} \quad
\texttt{aartisingh@cmu.edu}
}
\hypersetup{pdftitle={Joint Constraint Aggregation for Qualifier-Sensitive Retrieval-Augmented QA},
            pdfauthor={Jennifer Hsia, Zachary C. Lipton, Aarti Singh}}
\begin{document}
\maketitle

\begin{abstract}
Retrieval-augmented QA can fail even when the correct answer is in the retrieved evidence. This can happen when a competing answer satisfies some, but not all, of the question’s requirements and is mentioned frequently in the retrieved documents, which can make it appear to be the best-supported answer. To test whether this retrieval skew actually affects the decision, we manipulate it directly: increasing the prevalence of passages that favor the competing answer makes the model more likely to select it. This points to a decision problem that candidate-level support alone cannot resolve. We therefore introduce Joint Constraint Aggregation (JCA), which compares candidates against the same question-derived requirements using the retrieved evidence. Across three evaluation settings and three instruction-tuned backbones, JCA improves overall accuracy, and its advantage grows when retrieval favors a competing answer. Together, these results suggest that when partial matches and retrieval prominence pull models toward a competing answer, comparing candidates against the question’s full requirements can make selection more reliable.
\end{abstract}

\section{Introduction}

Retrieval-augmented QA \citep{lewis2020retrieval,izacard2021leveraging} can
fail even when the retrieved evidence contains the correct answer. One difficult case arises when a competing answer is also
plausible: it may be correct for a closely related interpretation of the
question, but not for the question as stated. A time period, version, role,
jurisdiction, or other qualifier may determine which otherwise-supported
answer applies. We study this problem as \emph{qualifier-sensitive answer
selection}: given retrieved evidence and a set of plausible candidates that
includes the correct answer, select the candidate that answers the question
as stated. Conditioning on the correct candidate being available isolates
selection from failures to retrieve or generate it.

\begin{figure*}[t]
    \centering
    \includegraphics[width=0.99\linewidth]{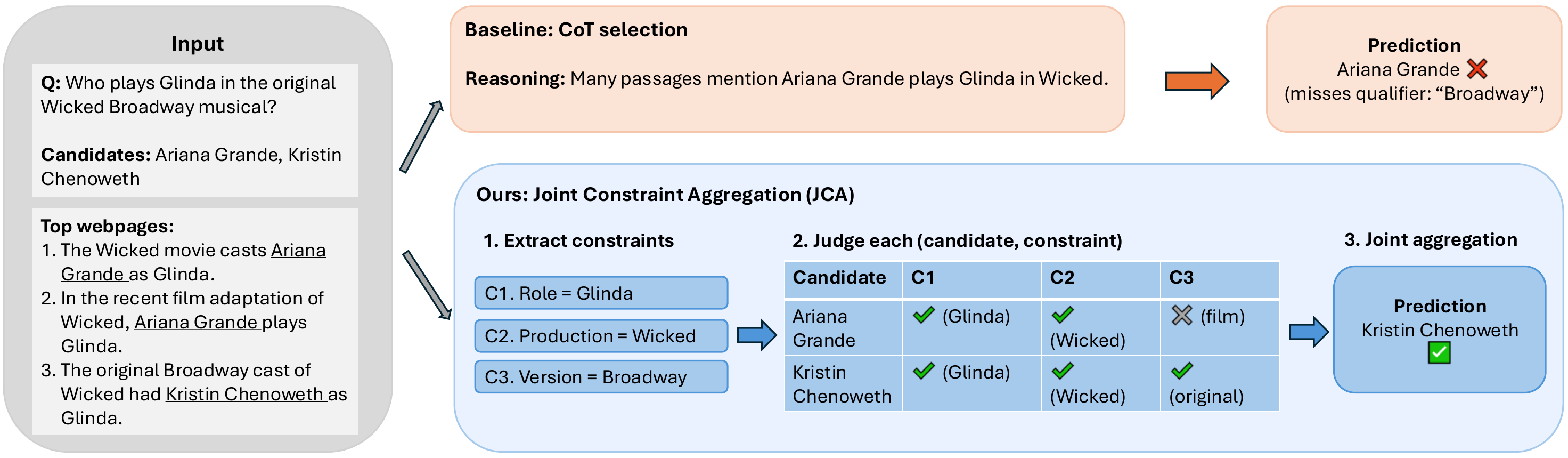}
    \caption{
Illustration of qualifier-sensitive answer selection and Joint Constraint
Aggregation. Retrieved webpages mention a plausible near-miss, Ariana
Grande in the film adaptation of \textit{Wicked}, more frequently than the
qualifier-satisfying answer, Kristin Chenoweth in the original Broadway
production. A baseline selector chooses the near-miss despite the Broadway
qualifier. JCA represents the question requirements as shared constraints,
checks each candidate against those constraints using retrieved evidence, and
jointly compares the resulting reports.
}
    \label{fig:jca}
\end{figure*}

The difficulty is not simply whether each candidate has supporting evidence.
Evidence that supports a candidate is not necessarily evidence that the
candidate answers the question being asked. When several candidates have
legitimate support, the key distinction may be whether that evidence
establishes the particular requirement that makes one candidate applicable.
This motivates comparing candidates against the same question-derived
requirements rather than treating candidate-level support as sufficient for
selection. Figure~\ref{fig:jca} illustrates this setting and the
structured comparison we introduce below.

Retrieval can amplify this problem when it makes a plausible near-miss more
salient than the applicable answer. In naturally retrieved evidence, baseline
accuracy declines as passages mention the near-miss more often relative to
the correct answer. We
also manipulate this imbalance directly: holding the question, candidate set,
and neutral evidence fixed, shifting relative support toward the near-miss
increases near-miss selection. Together, these analyses show that relative
retrieval support can affect the downstream choice among plausible answers.

We introduce \textbf{Joint Constraint Aggregation (JCA)}\footnote{Code repository: \url{https://github.com/jenhsia/joint_constraint_aggregation}} to make comparison
under shared question requirements explicit. JCA represents the question
requirements as a shared set of atomic constraints, checks each candidate
against those same constraints using the retrieved evidence, and jointly
compares the resulting evidence-grounded reports. In this way, all candidates
are evaluated under the same question-derived criteria.

We evaluate JCA across three qualifier-sensitive settings spanning
human-annotated near-misses, candidates extracted from retrieved passages, and
temporal qualifiers. JCA improves selection on all three full evaluation
sets. On hard slices where retrieval mentions a plausible near-miss more often
than the correct answer, JCA achieves the highest accuracy in all nine
dataset--backbone combinations across three instruction-tuned LLM backbones.
Controlled analyses further establish the effect of relative retrieval
support and examine which parts of the structured decision procedure
contribute to JCA's gains.

\paragraph{Contributions.}
Our main contributions are:
\begin{itemize}
    \item We isolate qualifier-sensitive answer selection, where plausible
    alternatives may each have supporting evidence but differ in whether they
    satisfy the question as stated, and show through natural and controlled
    analyses that relative retrieval support can shift selection.

    \item We introduce JCA, which evaluates candidates against shared
    question-derived requirements and jointly compares their
    evidence-grounded reports.

    \item We demonstrate consistent gains across three qualifier-sensitive
    settings and three instruction-tuned LLM backbones, and use component and
    design analyses to characterize the structured decision procedure.
\end{itemize}

\section{Related Work}

\paragraph{Evidence support does not always determine applicability.}
Verification methods make evidence support explicit. In QA, predictions can
be recast as hypotheses and checked against retrieved context with natural
language inference \citep{chen2021can}, while factuality evaluation can
decompose generated text into atomic claims and verify each against external
evidence \citep{min2023factscore}. These approaches are closely related to
JCA's candidate-checking stage, which asks what the available evidence
establishes about each candidate.

Candidate reranking instead addresses selection among alternatives.
Open-domain QA rerankers aggregate evidence across passages
\citep{wang2017evidence} or learn to distinguish close false positives within
a candidate set \citep{iyer2021reconsider}. In our setting, however, multiple
candidates may each have legitimate supporting evidence. The remaining
decision is which candidate's evidence establishes the requirements imposed by
the question. JCA makes those requirements shared across candidates before
comparison.

\paragraph{Decomposing a problem differs from comparing candidates.}
Prompting methods can break a complex problem into simpler subproblems that
are solved in sequence, with each answer informing the next
\citep{zhou2022least}, or delegate subtasks to prompts specialized for them
\citep{khot2022decomposed}. These methods decompose the procedure for
producing an answer. JCA instead decomposes what the question requires and
applies the same requirements to every candidate, so that candidates with
partial support can be compared on equal terms. Decomposition alone also does
not explain JCA's gains. Our controls that use the same extracted requirements
but score candidates independently (Constrained NLI) or combine verdicts with a
fixed rule (Constraint Tally) remain below full JCA in every setting
(Table~\ref{tab:main_results} and Appendix~\ref{app:cross_backbone}).

\paragraph{Retrieval composition can shift candidate selection.}
Prior work shows that downstream QA depends on what retrieval surfaces.
Retrievers can favor more prominent entities under ambiguous queries
\citep{chen2021evaluating}, related but irrelevant passages can distract RAG
models \citep{amiraz2025distracting}, and imperfect retrieval can introduce
conflicting evidence \citep{wang2025astute}. Our setting complements these
findings: evidence favoring a near-miss need not be irrelevant, erroneous, or
contradictory, but may correctly support that candidate for a nearby
interpretation. We study the downstream choice when retrieval makes such a
plausible near-miss especially salient.

\paragraph{Ambiguity and context create plausible alternatives.}
Ambiguity-aware QA makes multiple valid interpretations of a question
explicit. AmbigQA pairs plausible answers with disambiguated question rewrites
\citep{min2020ambigqa}, while ASQA produces answers spanning multiple valid
readings \citep{stelmakh2022asqa}. AmbigDocs extends this ambiguity to
retrieved documents. When documents describe different entities that share a
name, models often merge information across them \citep{lee2024ambigdocs}.
There, the question itself is ambiguous and a complete answer covers each
entity. In our setting, the question specifies which reading applies, and
evidence for the near-miss can be accurate while answering a different
reading. These settings illustrate why a competing
answer can be genuinely plausible rather than simply false: it may answer a
related interpretation of the question.

Answer validity can also depend on specified context. SituatedQA studies
questions whose answers vary with temporal or geographic context
\citep{zhang2021situatedqa}. Our evaluation draws on both forms of variation
but focuses on the downstream decision among plausible alternatives once the
target interpretation or context is specified.

\section{Experimental Setup}
\label{sec:setup}
\subsection{Task and Evaluation Design}

We study retrieval-augmented QA as candidate selection. Each instance contains
a question, retrieved passages, and a small set of plausible candidate answers
that includes the correct answer. The task is to select the candidate that
answers the question as stated. We therefore evaluate selection conditional on
candidate recall rather than failures to retrieve, extract, or generate the
correct candidate.

We use three evaluation conditions that vary where the candidate alternatives
come from and what determines which candidate applies. AmbigQA uses
human-annotated alternatives associated with nearby interpretations.
AmbigQA-Extracted keeps the ambiguity setting but derives candidates from
retrieved passages. SituatedQA-temporal instead tests applicability determined
by explicit temporal context.

\paragraph{AmbigQA.}
We begin with AmbigQA \citep{min2020ambigqa}, which provides disambiguated question–answer pairs
for questions with multiple valid interpretations. We use its human-annotated
alternatives to form candidate sets in which a distractor can be a valid answer
to a nearby interpretation even though it does not answer the target question.
This condition therefore tests selection among plausible alternatives supplied
by the dataset.

\paragraph{AmbigQA-Extracted.}
To test the same type of selection when candidates come from retrieval rather
than human annotation, we construct AmbigQA-Extracted from AmbigQA training
questions. Qwen3-30B extracts plausible candidates from the retrieved passages
without access to the gold answer. The resulting candidate sets are generated
once and fixed across all methods and evaluated backbones. After
deduplication, we evaluate only instances in which the gold answer is recovered
and at least two distinct candidate groups remain.
Appendix~\ref{app:candidate_extraction} reports
candidate-recovery coverage, and the code release documents the full extraction procedure.

\paragraph{SituatedQA-temporal.}
We use the temporal subset of SituatedQA \citep{zhang2021situatedqa} to test a
different source of applicability. The same question can have different
correct answers at different times. For each instance, the answer associated
with the target temporal context is the gold candidate, while answers that are
correct for the same question under other temporal contexts serve as plausible
distractors. This condition therefore tests whether the model selects the answer that applies to the time specified by the question.

\subsection{Evidence and Instance Construction}

For all three conditions, we retrieve the top-$k=20$ results for each question
using the Brave Search API\footnote{\url{https://brave.com/search/api/}} and use their extracted page text as evidence. All
methods begin from the same ordered top-20 passage pool. Passage text is
truncated as needed to fit the context available to each backbone. Appendix~\ref{app:retrieval}
provides retrieval, mention-matching, and truncation details.

\paragraph{Instance eligibility.}
To make selection errors interpretable, an instance must present a
well-defined choice among alternatives that are actually surfaced by the
evidence. For AmbigQA and SituatedQA-temporal, we therefore require the
correct candidate and at least one distractor to each be mentioned in at
least 10\% of the top-20 retrieved passages, using alias-based whole-word
matching. We do not apply this requirement to AmbigQA-Extracted because its
candidates are themselves extracted from the retrieved passages. We also
exclude malformed instances or answer structures that do not support a
well-defined single-candidate selection problem. Exact screening criteria and
counts are provided in the code release.

For a selection error to be meaningful, the gold candidate must also be
uniquely applicable to the target question. Because the alternatives are
deliberately drawn from nearby valid interpretations or contexts, a nominal
distractor can occasionally also be a defensible answer to the exact target
question. To avoid counting such a choice as an error, we use GPT-5.5 \citep{openai2026gpt55systemcard} to
classify each distractor as \textsc{distinct}, \textsc{co-correct}, or
\textsc{unsure}. We exclude an instance if any distractor is judged
\textsc{co-correct} and retain cases judged \textsc{distinct} or
\textsc{unsure}. Appendix~\ref{app:cocorrect} reports filtering counts and an author audit.
The exact judge prompt and filtering implementation are provided in the code
release.

\subsection{Models and Evaluation}

We evaluate three instruction-tuned LLM backbones from different model
families: Llama4-Scout-17B \citep{meta2025llama4}, Qwen3-30B \citep{yang2025qwen3}, and Gemma3-27B \citep{team2025gemma}. We use the checkpoints Qwen3-30B-A3B-Instruct-2507, gemma-3-27b-it, and a
4-bit quantized (w4a16) Llama-4-Scout-17B-16E-Instruct.
Section~\ref{sec:design-tradeoffs} separately examines Qwen3's reasoning
variant, Qwen3-30B-A3B-Thinking-2507, which we refer to as reasoning mode. Full implementation details are provided in the code
release.

We report candidate-selection accuracy on each full evaluation set and on a diagnostic hard slice where retrieval mention frequency favors a plausible near-miss. Unless noted otherwise, we compare methods with 95\% paired-bootstrap confidence intervals over instances \citep{efron1994introduction,koehn2004statistical} and call a difference statistically significant when its interval excludes zero.

\paragraph{Hard slice.}
\label{sec:hard_slice}
For each instance, we define the mention gap
\[
g = r(\text{gold}) - \max_{d \in D} r(d),
\]
where $r(\cdot)$ is the fraction of the fixed top-20 retrieved passages that
mention a candidate and $D$ is the set of distractors. The hard slice contains
instances with $g < 0$. Thus, at least one distractor is mentioned in more
retrieved passages than the gold answer. This yields 294 AmbigQA, 198 AmbigQA-Extracted, and 238 SituatedQA-temporal instances. We define slice membership from the fixed top-20 retrieval before
model-specific truncation. This keeps the evaluated slice identical across
backbones rather than allowing each model's context window to determine
membership. Appendix~\ref{app:hard_slice_sensitivity} reports a sensitivity analysis that recomputes the
gap from model-visible evidence.

Mention gap is a diagnostic proxy for retrieval salience. It does not measure
semantic support, and the frequency of hard-slice instances should not be
interpreted as an estimate of how often this problem occurs in deployed
systems.

\section{Methods}
\label{sec:methods}

We denote a question by $q$, retrieved passages by $P$, and candidate answers by
$A=\{a_1,\ldots,a_n\}$. Evidence construction, filtering, and model setup are
described in Section~\ref{sec:setup}.

\subsection{Comparison Methods}
\label{sec:baselines}

\paragraph{Set-level selection.}
Three baselines receive the question, retrieved passages, and candidate set
together and return one candidate in a single completion. \textbf{Direct}
selects an answer without an explicit reasoning instruction. \textbf{CoT} \citep{wei2022chain}
instead prompts the model to reason over the retrieved evidence before
selecting. \textbf{Comparative Prompting (Compare)} explicitly asks the model
to compare the candidates before choosing among them. These approaches vary
how reasoning and comparison are elicited, but do not explicitly represent the
question requirements as a shared constraint set with intermediate
candidate-level reports.

\paragraph{Independent evidence scoring.}
\textbf{NLI} scores each candidate separately. For each candidate $a_i$, we
form the hypothesis ``The answer to the question $q$ is $a_i$'' and score it
against $P$. Following \citet{chen2023menli} and \citet{steen2023little}, we
compute
\begin{equation}
    \operatorname{score}(a_i)
    = P(\textsc{Entail}) - P(\textsc{Contradict}),
\end{equation}
where $P(\cdot)$ is a normalized label probability read from the model's
output distribution at the label position. The candidate with the highest
score is selected.

We additionally evaluate two cross-encoder rerankers and CoT-SC, a
self-consistency variant of CoT \citep{wang2022self}. Their definitions and results are reported in
Appendices~\ref{app:reranker} and~\ref{app:self_consistency}.

\subsection{Joint Constraint Aggregation}
\label{sec:ours}

JCA makes comparison under shared question
requirements explicit. It represents the question requirements as atomic
constraints, checks each candidate against the same constraints using the
retrieved evidence, and then compares the resulting candidate reports jointly
(Figure~\ref{fig:jca}).

\paragraph{Step 1: Extract constraints.}
The model decomposes $q$ into a small set of $m$ atomic constraints. Each constraint captures one
requirement implied by the question, such as an answer type, role, time period,
version, or scope. This step operates on the question alone, so the same
constraint set is used for every candidate.

\paragraph{Step 2: Check candidates.}
For each candidate $a_i$, the model evaluates all $m$ constraints against the
retrieved passages $P$. For each constraint, it produces a verdict in
\{\textsc{Satisfies}, \textsc{Violates}, \textsc{Not\_Stated}\} and a brief
evidence-grounded rationale. Across candidates, this produces an $n\times m$
collection of reports under the same question requirements. In the primary
staged implementation, each candidate is checked in a separate model call.

\paragraph{Step 3: Jointly aggregate.}
The aggregation model receives the question, candidate set, and all
constraint-checking reports, including their verdicts and rationales. It
selects the candidate that best satisfies the question requirements relative
to the alternatives. Step~3 does not receive the retrieved passages directly,
so its decision is based on the information preserved in the candidate
reports, together with the model's own knowledge of the question and candidates.

The main experiments use staged JCA, with one constraint-extraction call,
one Step~2 call per candidate, and one final aggregation call. These three
operations define the JCA procedure, while their packaging into model calls
is an implementation choice examined below.

\subsection{Controls and Variants}
\label{sec:ablations}

\paragraph{Step~3 controls.}
To analyze the final aggregation while holding the earlier stages fixed,
\textbf{Constraint Tally} and \textbf{Verdict-only aggregation} both reuse the
staged JCA constraint checks. Constraint Tally replaces Step~3 with a
deterministic signed tally over the verdicts:
$\textsc{Satisfies}=+1$, $\textsc{Not\_Stated}=0$, and
$\textsc{Violates}=-1$. The candidate with the highest tally is selected, with
$1/k$ credit when the gold candidate is among $k$ tied winners.
Verdict-only aggregation instead removes the rationales from the same Step~2
reports before the joint Step~3 decision, leaving only the constraint verdict
labels. Comparing Constraint Tally with verdict-only aggregation varies the
decision rule while holding the verdict labels fixed. Comparing verdict-only
aggregation with full JCA varies rationale availability while holding the
joint aggregation procedure fixed.

\paragraph{Constraint-aware independent scoring.}
\textbf{Constrained NLI} provides a different control. It uses JCA's extracted
question constraints but does not reuse JCA's Step~2 reports. Instead, for
each candidate it jointly produces its own per-constraint judgments and an NLI
label from the retrieved evidence, then applies the same
entailment-minus-contradiction scoring rule as NLI. This retains independent
candidate scoring while making the question constraints explicit.

\paragraph{Call-structure variants.}
We also evaluate two variants that change how JCA's operations are packaged
into calls. \textbf{Merged-check JCA} retains the three-stage procedure but
checks all candidates together in a single Step~2 call rather than one call
per candidate. \textbf{OnePass JCA} instead asks a single completion to
extract the constraints, evaluate the candidates, and select the final answer.
These variants distinguish batching within the checking stage from collapsing
the explicit stage boundaries. Exact prompts and implementation details are provided in the code release.

\begin{table*}[ht]
\centering
\small
\setlength{\tabcolsep}{5pt}
\caption{
Accuracy (\%) on the full evaluation set and hard slice across three
evaluation conditions using Llama4-Scout-17B.
\textbf{Hard}: mention gap $<0$.
$n$ gives full / hard instances.
JCA achieves the highest accuracy in every column.
}
\label{tab:main_results}
\begin{tabular}{lcccccc}
\toprule
& \multicolumn{2}{c}{\textbf{AmbigQA}}
& \multicolumn{2}{c}{\textbf{AmbigQA-Extracted}}
& \multicolumn{2}{c}{\textbf{SituatedQA}} \\
& \multicolumn{2}{c}{($n$ = 796 / 294)}
& \multicolumn{2}{c}{($n$ = 548 / 198)}
& \multicolumn{2}{c}{($n$ = 674 / 238)} \\
\cmidrule(lr){2-3}\cmidrule(lr){4-5}\cmidrule(lr){6-7}
\textbf{Method}     & Full  & Hard  & Full  & Hard  & Full  & Hard  \\
\midrule
Direct              & 84.2  & 75.9  & 84.5  & 79.3  & 62.9  & 47.5  \\
CoT                 & 87.4  & 82.3  & 89.6  & 85.9  & 74.2  & 64.3  \\
Compare             & 82.9  & 71.1  & 81.0  & 68.7  & 65.9  & 47.1  \\
NLI                 & 78.9  & 66.7  & 83.0  & 77.3  & 59.6  & 45.4  \\
\midrule
\textbf{JCA} (ours)
                    & \textbf{93.3} & \textbf{92.5}
                    & \textbf{93.6} & \textbf{91.9}
                    & \textbf{85.5} & \textbf{79.8} \\
\midrule
Constrained NLI     & 83.3  & 76.2  & 85.6  & 84.3  & 78.3  & 73.9  \\
Constraint Tally    & 85.3  & 79.7  & 86.7  & 84.9  & 74.9  & 68.9  \\
\bottomrule
\end{tabular}

\end{table*}

\begin{table*}[ht]
\centering
\small
\setlength{\tabcolsep}{4pt}
\caption{
Hard-slice accuracy (\%) across instruction-tuned backbones.
Each cell reports JCA accuracy with the strongest primary baseline in
parentheses, where primary baselines are Direct, CoT, Compare, and NLI.
JCA achieves the highest hard-slice accuracy in all nine
dataset--backbone combinations.
Full comparison grids are reported in
Appendix~\ref{app:cross_backbone}.
}
\label{tab:backbone_hard_compact}
\begin{tabular}{lccc}
\toprule
\textbf{Backbone}
    & \textbf{AmbigQA}
    & \textbf{AmbigQA-Extracted}
    & \textbf{SituatedQA} \\
\midrule
Qwen3-30B
    & \textbf{93.5} (86.7)
    & \textbf{95.5} (88.9)
    & \textbf{81.1} (60.9) \\
Gemma3-27B
    & \textbf{79.3} (74.8)
    & \textbf{88.9} (81.8)
    & \textbf{63.4} (46.6) \\
Llama4-Scout-17B
    & \textbf{92.5} (82.3)
    & \textbf{91.9} (85.9)
    & \textbf{79.8} (64.3) \\
\bottomrule
\end{tabular}
\end{table*}
\section{Results}
\label{sec:results}

We begin by establishing JCA’s gains on the full evaluation sets and hard slices, then examine how relative retrieval support affects selection among plausible answers. We next hold the candidate checks fixed to isolate what the final joint aggregation contributes. Finally, we examine how the procedure behaves under alternative call structures and with a reasoning model.

\subsection{Main Selection Results}
\label{sec:main-results}

\paragraph{JCA improves selection beyond the targeted hard regime.}
We first ask whether JCA's gains are confined to the frequency-skewed cases
that motivate our diagnostic analysis. They are not. With
Llama4-Scout-17B, JCA achieves the highest accuracy on all three full
evaluation sets (Table~\ref{tab:main_results}), and the same holds for
Qwen3-30B and Gemma3-27B (Appendix~\ref{app:cross_backbone}). This establishes that the
advantage extends to qualifier-sensitive candidate selection more broadly,
rather than appearing only when retrieval favors a near-miss.

We next examine what changes in that more challenging regime. On the hard
slice, a plausible near-miss is mentioned in more retrieved passages than the
correct answer, making retrieval salience less aligned with the question.
Here the separation between JCA and the primary baselines becomes larger.
Against CoT, the strongest primary baseline with Llama4-Scout, JCA's margins
range from 4.0--11.3 points on the full sets and from 6.1--15.5 points on the
hard slices. Thus, JCA's advantage is present on the full evaluation, but is
more pronounced when retrieval favors a competing answer. The comparisons in
Table~\ref{tab:main_results} are statistically significant except for JCA
versus Constrained NLI on the SituatedQA hard slice, where JCA remains higher
but the confidence interval includes zero.

\paragraph{The hard-slice result holds across backbones.}
The Llama4 results leave open whether this pattern depends on a particular
model family. Table~\ref{tab:backbone_hard_compact} therefore extends the
hard-slice comparison to Qwen3-30B and Gemma3-27B. Across the three evaluation
conditions and three instruction-tuned backbones, JCA achieves the highest
hard-slice accuracy in all nine combinations. The strongest competing
baseline varies across settings, but the JCA ranking remains unchanged.

\subsection{Retrieval Support and Selection}
\label{sec:frequency-bias-analysis}

Section~\ref{sec:main-results} showed that differences between selection
methods become larger on the hard slice. We now ask a more basic question:
does what the model selects change as retrieval increasingly favors a
plausible near-miss?

\begin{figure}[ht]
    \centering
    \includegraphics[width=0.99\linewidth]{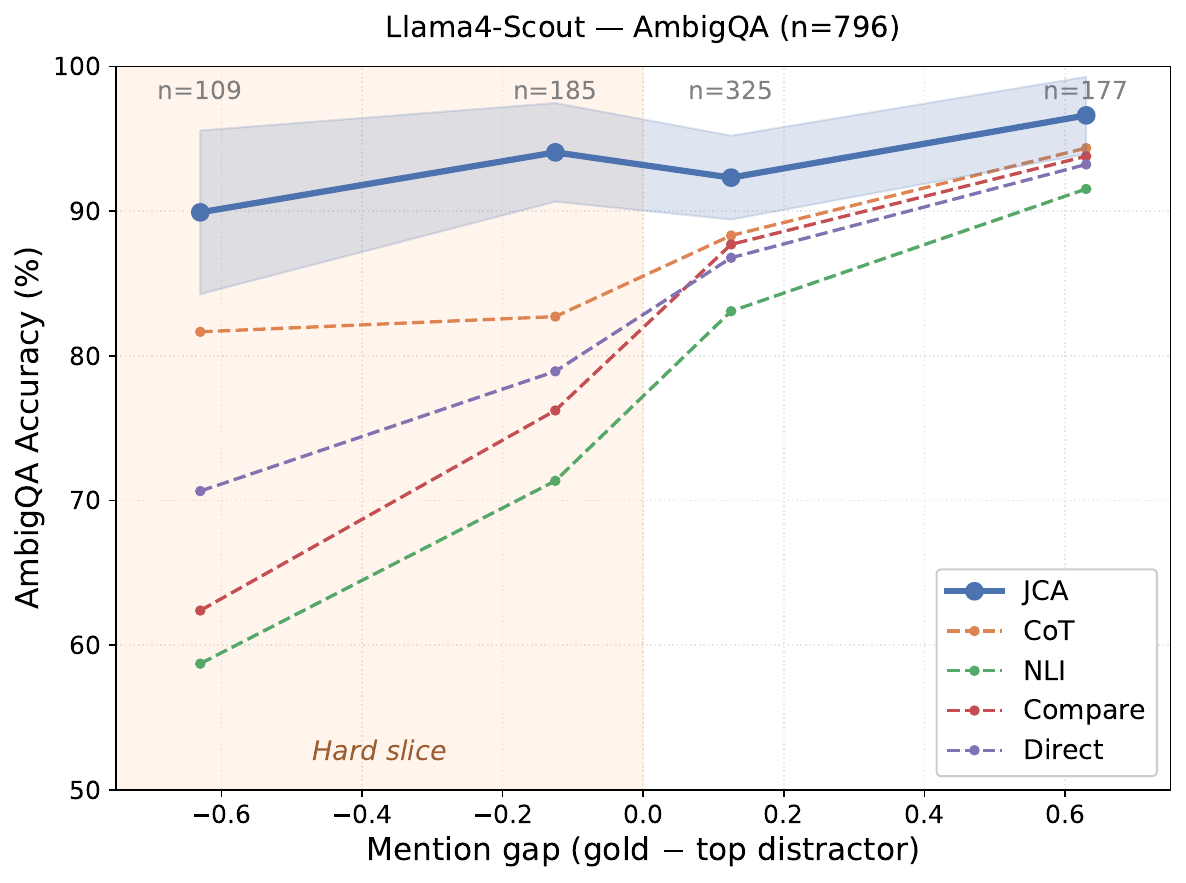}
    \caption{
    Accuracy versus mention gap on AmbigQA with Llama4-Scout-17B.
    More negative values indicate that retrieval mentions a plausible
    distractor more often relative to the gold answer. Baseline accuracy
    declines as the gap becomes more negative, while JCA remains
    comparatively stable. Shaded bands show 95\% confidence intervals for
    JCA.
    }
    \label{fig:llama_ambigqa_frequency_bias}
\end{figure}

\paragraph{Selection varies with relative retrieval support.}
Figure~\ref{fig:llama_ambigqa_frequency_bias} examines this relationship in
the naturally retrieved AmbigQA evidence. The mention gap measures how much
surface prominence retrieval gives the gold answer relative to the
most-mentioned distractor. As the gap becomes more negative, and retrieval
increasingly favors the near-miss, the baseline methods become less accurate.
JCA is comparatively stable across the same range. Importantly, mention
frequency is only a measure of retrieval salience: a frequently mentioned
distractor may be well supported for a related interpretation without
answering the question as stated.

\begin{table*}[ht]
\centering
\small
\setlength{\tabcolsep}{3.5pt}
\caption{
Contribution of JCA's final aggregation stage with Llama4-Scout-17B.
Panel A reports full-set accuracy for three aggregation rules applied to the
same Step~2 reports. Panel B shows the accuracy change from Constraint Tally
to full JCA within different Step~2 verdict outcomes, based on each candidate's Constraint Tally score.
In \emph{Decisive} cases the gold candidate scores strictly higher than every
distractor. In \emph{Tied} cases it shares the top score with at least one
distractor. In \emph{Gold outranked} cases a distractor outranks it and at
least one candidate has a positive score. Rare no-pass cases, where a
distractor outranks the gold candidate and no candidate has a positive score,
are omitted from Panel B. Differences in the
text are computed from unrounded accuracies.
}
\label{tab:step3}

\begin{tabular}{lccc}
\toprule
& AmbigQA & Extracted & SituatedQA \\
\midrule

\multicolumn{4}{l}{\textit{A. Full-set accuracy (\%)}} \\
Constraint Tally
    & 85.3 & 86.7 & 74.9 \\
Verdict-only aggregation
    & 89.1 & 90.5 & 84.1 \\
Full JCA
    & \textbf{93.3} & \textbf{93.6} & \textbf{85.5} \\

\midrule

\multicolumn{4}{l}{\textit{B. Step~3 gain by verdict outcome (pp)}} \\
Decisive
    & $-$0.5 & $-$0.7 & $-$3.2 \\
Tied
    & +35.5 & +34.8 & +32.2 \\
Gold outranked
    & +31.4 & +28.6 & +26.2 \\

\bottomrule
\end{tabular}
\end{table*}

This natural-data pattern is suggestive, but it does not by itself show that
retrieval support is causing the change in selection. Questions with more
negative mention gaps could simply be harder in other ways. We therefore
manipulate relative support directly while holding the question, candidate
set, and neutral evidence fixed.

When support is shifted from the gold answer toward the near-miss, the
model selects the near-miss more often. This endpoint increase occurs in all 15
dataset--method conditions tested, ranging from 1.4$\times$ to 9.7$\times$.
The fitted trends also point in the same frequency-sensitive direction in all
15 conditions (Qwen3-30B, Appendix~\ref{app:retrieval_intervention}). Together,
these results provide causal evidence, within our controlled evidence sets, that
changing relative retrieval support affects answer selection.

The intervention is designed to establish that phenomenon, not to rank the
methods by causal robustness. Most between-method slope
comparisons (20 of 24) are not statistically significant, so we do not infer
that one method is generally less sensitive to retrieval support than another.

\subsection{What Does Joint Aggregation Add?}
\label{sec:aggregation}

JCA's final stage compares the candidate reports produced by the shared
constraint checks. We ask whether that joint comparison adds anything once the
Step~2 reports already exist, and whether any additional benefit comes from
the rationale text in those reports. To separate these effects, we compare
three aggregation rules over the same Step~2 outputs: deterministic Constraint
Tally, joint aggregation over the verdict labels alone, and full JCA with the
rationales restored.

Table~\ref{tab:step3} shows that joint comparison helps even without access to
the rationales. Replacing Constraint Tally with verdict-only aggregation
improves full-set accuracy on all three evaluation conditions, by 3.8 points
on both AmbigQA conditions and 9.3 points on SituatedQA. Restoring the
rationales provides an additional 3--4 points on the two AmbigQA conditions,
while the 1.3-point increase on SituatedQA is not statistically significant.
The Step~3 gain therefore cannot be explained only by rationale content.
Joint comparison adds value beyond a fixed decision rule over the same
verdicts (although the Step~3 model can also draw on its own knowledge), while
rationales can preserve additional information beyond the verdict labels.

\begin{table*}[ht]
\centering
\small
\setlength{\tabcolsep}{5pt}
\caption{
Hard-slice accuracy (\%) with Qwen3-30B under the tested instruction and
native-reasoning settings. OnePass collapses JCA's three operations into a
single completion, while staged JCA keeps them explicit. Paired differences
and 95\% confidence intervals are reported in Appendix~\ref{app:design_inference}.
}
\label{tab:design_reasoning}
\begin{tabular}{lccccc}
\toprule
& \multicolumn{3}{c}{\textbf{Instruction}}
& \multicolumn{2}{c}{\textbf{Reasoning}} \\
\cmidrule(lr){2-4}
\cmidrule(lr){5-6}
\textbf{Dataset}
& \textbf{OnePass JCA}
& \textbf{CoT}
& \textbf{Staged JCA}
& \textbf{CoT}
& \textbf{Staged JCA} \\
\midrule
AmbigQA
& 83.3 & 86.4 & 93.5
& 94.6 & 95.6 \\
AmbigQA-Extracted
& 85.9 & 86.9 & 95.5
& 90.4 & 91.4 \\
SituatedQA
& 63.4 & 60.5 & 81.1
& 87.4 & 90.3 \\
\bottomrule
\end{tabular}
\end{table*}

The benefit is concentrated in cases where the verdicts do not already settle
the choice. When the constraint checks decisively favor the gold candidate,
full JCA is slightly below Constraint Tally ($-$0.5 to $-$3.2 points). In contrast,
when multiple candidates are tied under the verdicts, the total gain from
Step~3 exceeds 30 points in all three conditions. Large gains also occur when
the verdicts rank a distractor above the gold answer. These cases are precisely
where a fixed tally leaves the final decision unresolved or points in the
wrong direction.

Rationale content can also matter when verdict labels do not distinguish the candidates. It contributes
little when the verdict pattern is already decisive, but can retain
qualifier-specific distinctions that are lost when each constraint check is
compressed to a categorical label. Appendix~\ref{app:qualitative_step3} gives an example in which the
verdict patterns alone do not distinguish the candidates, while the
accompanying rationales preserve the information needed for the final choice.

\subsection{Design and Inference Trade-offs}
\label{sec:design-tradeoffs}

We next vary how JCA's three operations are packaged into calls and how the
procedure interacts with Qwen3's reasoning mode.

\paragraph{Batching candidate checks differs from collapsing the procedure.}
OnePass JCA performs constraint extraction, candidate checking, and final
selection within one completion. In the tested instruction-model settings,
this substantially reduces accuracy: across Qwen3-30B and Llama4-Scout-17B,
OnePass trails staged JCA by approximately 4--10 points on the full evaluation
sets and 10--18 points on the hard slices, with all differences statistically
significant.

By contrast, Merged-check JCA retains the three operations but evaluates all
candidates together in Step~2. Pooled over the 2,018 full-set instances within
each backbone, Merged-check is non-inferior to staged JCA at the prespecified
2.5-point margin for Qwen3-30B, Gemma3-27B, and Llama4-Scout-17B, with
accuracy differences of $+0.74$, $+0.05$, and $+0.25$ points, respectively.
Thus, separate per-candidate Step~2 calls are not required to preserve pooled
accuracy, whereas collapsing the full procedure into one completion degrades
accuracy in the tested instruction settings.

\paragraph{Merged and staged JCA occupy different inference operating points.}
In a dedicated Qwen3-30B measurement pass, Merged-check reduces the measured
candidate-checking and answer-stage input by 55\% on AmbigQA and 49\% on
SituatedQA relative to staged JCA, while also using fewer requests. Staged
JCA, however, exposes independent candidate checks that can be dispatched
concurrently. Under our derived candidate-parallel accounting, it has lower
latency on both measured datasets. Thus, batching reduces repeated input and
request count, while separate checks can favor parallel latency. Full
measurements and accounting details are reported in
Appendix~\ref{app:cost_measurement}.

\paragraph{The staged-JCA advantage is much smaller with reasoning enabled.}
Without Qwen3's reasoning mode, staged JCA exceeds CoT by 7.1--20.6 points
across the three hard slices. With reasoning enabled for both methods, the
differences shrink to $+1.0$, $+1.0$, and $+2.9$ points (Table~\ref{tab:design_reasoning}),
and none is statistically significant. We therefore do not establish an
additional accuracy benefit for staged JCA in this Qwen3 reasoning setting.
At the prespecified 2.5-point margin, staged JCA with reasoning is
non-inferior to CoT with reasoning on AmbigQA and SituatedQA, while the
AmbigQA-Extracted comparison is too imprecise to certify
(Appendix~\ref{app:reasoning_mode}).

Overall, the results separate the structured procedure from its most expensive
implementation: collapsing all three operations hurts in the tested
instruction settings, whereas batching candidate checks preserves pooled
accuracy while substantially reducing repeated input.

\section{Conclusion}
\label{sec:conclusion}

A candidate can be well supported by retrieved evidence and still be the wrong answer to the question being asked. Our experiments show that this distinction matters when several plausible answers are supported: retrieval can make a near-miss more salient, while explicitly comparing candidates under the same question requirements substantially improves selection in the tested instruction settings. The relevant question is therefore not only whether evidence supports an answer, but whether it establishes that the answer applies to the question at hand.

Across the component and design analyses, a consistent picture emerges. The useful abstraction is comparison under shared question requirements, rather than one separate call per candidate. Collapsing all three stages into a single completion, however, substantially hurts accuracy. Joint comparison adds value beyond fixed aggregation, while candidate checks can be packaged differently without giving up the structured decision process. JCA therefore provides one concrete way to preserve the distinction between evidence that supports an answer and evidence that establishes that the answer applies to the question at hand.

\section*{Limitations}
\label{sec:limitations}

Our experiments isolate answer selection conditional on candidate recall.
They therefore do not measure whether an end-to-end RAG system retrieves or
generates the correct candidate in the first place. This distinction is
substantial in AmbigQA-Extracted: the extraction pipeline recovers the gold
answer for 68.6\% of questions that produce candidates, and candidate recovery therefore remains a substantial upstream bottleneck. JCA improves the subsequent
selection decision but does not address this recovery bottleneck.

Our evaluation also prioritizes clean, distinguishable near-misses. The
LLM-assisted co-correct filter removes some valid cases along with ambiguous
ones, and the hard slice is defined by alias-based mention frequency rather
than semantic support. We therefore treat it as a diagnostic slice, not an
estimate of how often such failures occur in deployed systems. Its membership
can also change when the mention gap is recomputed over only the evidence
visible within a model's context window.

Finally, the accuracy benefit of explicit JCA structure depends on the model
regime. Our main gains are obtained with instruction-tuned models. With
Qwen3-30B's reasoning variant, staged JCA remains 1--3 points
higher than CoT-thinking on the three hard slices, but none of these
differences is statistically significant. Whether explicit JCA provides
additional accuracy benefits for other reasoning models or substantially
larger candidate sets remains open.

Staged JCA also costs more than single-call baselines: it issues one call per
candidate plus extraction and aggregation calls, although merged candidate
checking substantially reduces repeated input
(Appendix~\ref{app:cost_measurement}). Our evidence comes from a snapshot of live web search. We release the
retrieved URLs, titles, snippets, and per-candidate mention statistics, but not
the page text, and pages fetched later may differ from the originals.

\section*{Acknowledgments}
This work was supported in part by the NSF--USDA AIIRA grant
2021-67021-35329 and the Simons Foundation grant 888970.

\bibliography{references}

\appendix

\section{Experimental and Evaluation Details}
\label{app:experimental_details}

\subsection{Retrieval and Mention Matching}
\label{app:retrieval}

For each question, we retrieve the top-$k{=}20$ results from the Brave Search
API and use the extracted page text as evidence. All methods begin from the
same ordered passage pool. Retrieved text is limited to the smaller of a
48k-token evidence cap and the context space available to the backbone. The
48k cap is operative for Qwen3-30B and Gemma3-27B. For
Llama4-Scout-17B, the effective evidence budget is approximately 31.5k tokens,
with small method-dependent variation from prompt and output budgets.

Mention frequency is computed from the fixed top-20 retrieval before
model-specific truncation. We use alias-based whole-word matching, counting a
passage once for a candidate if any of its aliases appears. The mention rate
is the fraction of retrieved passages containing the candidate. Full retrieval,
matching, and truncation implementations are provided in the code release.

\subsection{AmbigQA-Extracted Candidate Recovery}
\label{app:candidate_extraction}

For AmbigQA-Extracted, we attempted candidate extraction on 2{,}000 AmbigQA
training questions using Qwen3-30B without access to the gold answer. At least
one candidate was extracted for 1{,}964 questions. Among these, the gold answer
was recoverable after extraction and deduplication for 1{,}347 questions
(68.6\%). Selection evaluation is then restricted to instances where the gold
answer is available and at least two distinct candidate groups remain, followed
by the evaluation filters described in Section~\ref{sec:setup}. The final
evaluation set contains 548 instances. Full extraction and deduplication
procedures are provided in the code release.

\subsection{Co-correct Filtering}
\label{app:cocorrect}

The co-correct filter removes instances where a labeled distractor is also a
defensible answer to the question as stated. For the final evaluation, we
apply a GPT-5.5 judgment pass to the eligible instances (Table~\ref{tab:cocorrect_counts}). The judge
receives only the question, gold answer, and distractor, without retrieved
passages or method outputs. It labels each distractor as
\textsc{distinct}, \textsc{co-correct}, or \textsc{unsure}. We remove an
instance if any distractor is labeled \textsc{co-correct}, and we retain
\textsc{distinct} and \textsc{unsure} cases.

\begin{table}[h]
\centering
\small
\setlength{\tabcolsep}{4pt}
\caption{Co-correct filtering counts for the final GPT-5.5 judgment pass.}
\label{tab:cocorrect_counts}
\begin{tabular}{lrrr}
\toprule
Stage & AmbigQA & Extracted & SituatedQA \\
\midrule
Before filtering & 875 & 715 & 792 \\
Removed                & 79  & 167 & 118 \\
Final evaluation set   & \textbf{796} & \textbf{548} & \textbf{674} \\
\bottomrule
\end{tabular}
\end{table}

\paragraph{Author audit.}
We manually audited a stratified sample of 75 judge decisions. Among 53 audited
triples retained by the filter, the author identified one clear co-correct case
(1.9\%). Among 22 audited triples removed by the filter, 8 were judged
co-correct, 9 distinct, and 5 unsure after lookup-based verification. Thus,
the filter reduces clear co-correct contamination while also removing some
valid near-miss cases. The exact judge prompt and filtering implementation are
provided in the code release.

\subsection{Sensitivity of the Hard-Slice Definition}
\label{app:hard_slice_sensitivity}

The primary hard slice is defined from the fixed top-20 retrieval so that the
same instances are evaluated across backbones. As a sensitivity check, we
recompute the mention gap using only model-visible evidence. On AmbigQA, this
reduces the hard slice from 294 instances to 248--260 depending on the
backbone. On SituatedQA, it reduces the slice from 238 to 157--185. We
therefore treat the fixed-top-20 hard slice as a backbone-independent
diagnostic definition rather than a claim about the exact salience distribution
inside every model's visible context.

We do not report the corresponding AmbigQA-Extracted sensitivity because the
available model-visible reconstruction covers only 480 of the 548 final
instances, rather than the complete evaluation set.

\section{Full Cross-Backbone Results}
\label{app:cross_backbone}

Table~\ref{tab:cross_backbone_full} expands the compact cross-backbone
comparison in Table~\ref{tab:backbone_hard_compact} to all comparison methods
and controls. JCA achieves the highest hard-slice accuracy in all nine
dataset--backbone combinations. Table~\ref{tab:cross_backbone_fullset} reports
the corresponding full-set accuracies, where JCA is likewise the most accurate
method in all nine combinations.

\begin{table*}[t]
\centering
\small
\setlength{\tabcolsep}{6pt}
\caption{Full-set accuracy (\%) across all three backbones.
Evaluation-set sizes are 796 for AmbigQA, 548 for AmbigQA-Extracted, and 674 for
SituatedQA. Best accuracy in each column is bolded.}
\label{tab:cross_backbone_fullset}
\begin{tabular}{lccc}
\toprule
\textbf{Method} & \textbf{Qwen3-30B} & \textbf{Gemma3-27B} & \textbf{Llama4-Scout-17B} \\
\midrule
\multicolumn{4}{l}{\textit{AmbigQA} ($n=796$)} \\
Direct & 88.3 & 66.0 & 84.2 \\
CoT & 90.7 & 79.6 & 87.4 \\
Compare & 91.7 & 74.4 & 82.9 \\
NLI & 90.8 & 83.9 & 78.9 \\
\textbf{JCA} & \textbf{94.0} & \textbf{87.4} & \textbf{93.3} \\
Constrained NLI & 79.4 & 68.0 & 83.3 \\
Constraint Tally & 87.3 & 77.9 & 85.3 \\
\midrule
\multicolumn{4}{l}{\textit{AmbigQA-Extracted} ($n=548$)} \\
Direct & 85.6 & 66.2 & 84.5 \\
CoT & 90.1 & 77.0 & 89.6 \\
Compare & 90.3 & 71.2 & 81.0 \\
NLI & 92.0 & 86.7 & 83.0 \\
\textbf{JCA} & \textbf{95.1} & \textbf{88.7} & \textbf{93.6} \\
Constrained NLI & 83.8 & 65.1 & 85.6 \\
Constraint Tally & 88.6 & 78.6 & 86.7 \\
\midrule
\multicolumn{4}{l}{\textit{SituatedQA} ($n=674$)} \\
Direct & 56.5 & 55.6 & 62.9 \\
CoT & 76.1 & 64.7 & 74.2 \\
Compare & 75.2 & 59.5 & 65.9 \\
NLI & 74.3 & 66.8 & 59.6 \\
\textbf{JCA} & \textbf{85.0} & \textbf{75.7} & \textbf{85.5} \\
Constrained NLI & 72.4 & 66.5 & 78.3 \\
Constraint Tally & 74.0 & 66.8 & 74.9 \\
\bottomrule
\end{tabular}
\end{table*}

\begin{table*}[t]
\centering
\small
\setlength{\tabcolsep}{6pt}
\caption{Hard-slice accuracy (\%) across all three backbones.
Hard-slice sizes are 294 for AmbigQA, 198 for AmbigQA-Extracted, and 238 for
SituatedQA. Slice membership is defined from the fixed top-20 retrieval and is
identical across backbones. Best accuracy in each column is bolded.}
\label{tab:cross_backbone_full}
\begin{tabular}{lccc}
\toprule
\textbf{Method}
& \textbf{Qwen3-30B}
& \textbf{Gemma3-27B}
& \textbf{Llama4-Scout-17B} \\
\midrule

\multicolumn{4}{l}{\textit{AmbigQA} ($n=294$)} \\
Direct            & 82.3 & 43.5 & 75.9 \\
CoT               & 86.4 & 67.7 & 82.3 \\
Compare           & 86.7 & 56.8 & 71.1 \\
NLI               & 85.7 & 74.8 & 66.7 \\
\textbf{JCA}      & \textbf{93.5} & \textbf{79.3} & \textbf{92.5} \\
Constrained NLI   & 75.9 & 63.3 & 76.2 \\
Constraint Tally  & 83.4 & 71.2 & 79.7 \\
\midrule

\multicolumn{4}{l}{\textit{AmbigQA-Extracted} ($n=198$)} \\
Direct            & 80.8 & 48.0 & 79.3 \\
CoT               & 86.9 & 70.2 & 85.9 \\
Compare           & 87.4 & 58.6 & 68.7 \\
NLI               & 88.9 & 81.8 & 77.3 \\
\textbf{JCA}      & \textbf{95.5} & \textbf{88.9} & \textbf{91.9} \\
Constrained NLI   & 82.8 & 67.7 & 84.3 \\
Constraint Tally  & 89.2 & 78.5 & 84.9 \\
\midrule

\multicolumn{4}{l}{\textit{SituatedQA} ($n=238$)} \\
Direct            & 32.4 & 32.4 & 47.5 \\
CoT               & 60.5 & 46.6 & 64.3 \\
Compare           & 60.9 & 36.1 & 47.1 \\
NLI               & 59.7 & 45.4 & 45.4 \\
\textbf{JCA}      & \textbf{81.1} & \textbf{63.4} & \textbf{79.8} \\
Constrained NLI   & 62.2 & 59.2 & 73.9 \\
Constraint Tally  & 70.2 & 54.0 & 68.9 \\
\bottomrule
\end{tabular}
\end{table*}

\section{Controlled Retrieval-Support Intervention}
\label{app:retrieval_intervention}

To test whether relative retrieval support itself can change answer selection,
we construct controlled evidence sets from each question's own top-20
retrieval and evaluate Qwen3-30B on them. For each instance, we
hold the question, candidate set, and neutral evidence fixed while varying the
ratio of passages supporting the gold answer to passages supporting a
near-miss. AmbigQA and AmbigQA-Extracted use gold:near-miss support ratios
$\{4{:}1, 2{:}1, 1{:}1, 1{:}2, 1{:}4\}$. The primary
SituatedQA analysis uses $\{2{:}1, 1{:}1, 1{:}2\}$ because its more extreme cells
did not meet the minimum distinct-instance threshold for the primary panel.
The primary outcome is whether the manipulated near-miss is selected.
Table~\ref{tab:retrieval_intervention} reports near-miss selection at the
endpoints of each support range.

\begin{table*}[t]
\centering
\small
\setlength{\tabcolsep}{3.5pt}
\caption{Near-miss selection at the endpoints of the controlled
retrieval-support intervention. ``Gold-fav.'' and ``Near-miss-fav.'' denote
the gold-favored and near-miss-favored endpoints, respectively. AmbigQA and
AmbigQA-Extracted compare $4{:}1$ with $1{:}4$ support, and SituatedQA compares
$2{:}1$ with $1{:}2$. The final column within each dataset gives the
multiplicative increase in near-miss selection.}
\label{tab:retrieval_intervention}
\begin{tabular}{lccc ccc ccc}
\toprule
& \multicolumn{3}{c}{\textbf{AmbigQA}}
& \multicolumn{3}{c}{\textbf{AmbigQA-Extracted}}
& \multicolumn{3}{c}{\textbf{SituatedQA}} \\
\cmidrule(lr){2-4}
\cmidrule(lr){5-7}
\cmidrule(lr){8-10}
\textbf{Method}
& Gold-fav. & Near-miss-fav. & $\times$
& Gold-fav. & Near-miss-fav. & $\times$
& Gold-fav. & Near-miss-fav. & $\times$ \\
\midrule
CoT
& 9.9\% & 26.7\% & 2.7
& 10.4\% & 33.3\% & 3.2
& 18.1\% & 32.5\% & 1.8 \\

CoT-thinking
& 3.6\% & 17.3\% & 4.8
& 2.1\% & 20.1\% & 9.7
& 9.2\% & 20.3\% & 2.2 \\

OnePass JCA
& 6.3\% & 20.0\% & 3.2
& 5.6\% & 35.2\% & 6.3
& 16.1\% & 33.3\% & 2.1 \\

Staged JCA
& 8.1\% & 16.0\% & 2.0
& 9.0\% & 19.5\% & 2.2
& 17.3\% & 23.8\% & 1.4 \\

Merged-check JCA
& 4.5\% & 16.0\% & 3.6
& 5.6\% & 26.6\% & 4.7
& 12.5\% & 24.7\% & 2.0 \\
\bottomrule
\end{tabular}
\end{table*}

Near-miss selection increases between the intervention endpoints in all 15
dataset--method conditions, by 1.4--9.7$\times$. The fitted analysis gives the
same directional result. We define
$x=\log(\text{gold support}/\text{near-miss support})$ and
$y=1$ when the near-miss is selected, so a negative coefficient indicates
greater near-miss selection as relative support shifts toward the near-miss.
The fitted coefficient is negative in all 15 conditions, although only 13 of
15 are individually significant. The endpoint result should not be read as
strict monotonicity over every intermediate support level: near-miss selection
is stepwise monotonic in 9 of 15 conditions.

The intervention fixes the neutral evidence, not the total amount of support
evidence. Mean realized input volume varies by about 1.77$\times$ across
support conditions. A sensitivity analysis therefore refits
the model with the designed support-slot count, $S_{\text{cell}}$, as an
additional covariate. The fitted retrieval-support coefficient remains within
its original bootstrap interval in all 15 conditions. This check supports the
interpretation that the directional effect is not explained by the associated
change in support volume.

Finally, the intervention tests whether changing relative retrieval support
affects selection, not whether one method is causally more robust than another.
Of 24 paired between-method slope comparisons, 20 are not
statistically significant. We therefore draw no causal ranking of the methods
from their fitted slopes.

\section{Example of Information Preserved by Rationale Content}
\label{app:qualitative_step3}

Consider the AmbigQA question \emph{``Which actress said `never mind' on
Saturday Night Live?''} Both Gilda Radner and Emily Litella receive the same
Step~2 verdict pattern, but the accompanying reports retain the
actress--character distinction needed by the question
(Table~\ref{tab:qual_step3}).

\begin{table*}[t]
\centering
\small
\setlength{\tabcolsep}{5pt}
\caption{Illustrative tied-verdict case. The Step~2 verdict patterns are
identical, while the report content preserves the distinction between the
actress and the character she played. Report content is abridged for
readability. Full JCA selects the gold answer, Gilda Radner.}
\label{tab:qual_step3}
\begin{tabular}{p{0.19\linewidth}p{0.19\linewidth}p{0.52\linewidth}}
\toprule
\textbf{Candidate} & \textbf{Verdict pattern} & \textbf{Abridged report content} \\
\midrule
\textbf{Gilda Radner} (gold)
& All \textsc{Satisfies}
& Gilda Radner played Emily Litella, the character who repeatedly said
``Never mind'' after being corrected. \\
\midrule
Emily Litella
& All \textsc{Satisfies}
& Emily Litella is the \emph{character} on \textit{Saturday Night Live} who
says ``Never mind.'' \\
\bottomrule
\end{tabular}
\end{table*}

A deterministic tally over the verdicts assigns the two candidates the same
score. The report text, however, preserves the relational fact that Gilda
Radner is the actress while Emily Litella is the character she played. Full
JCA selects Gilda Radner. This is a case in which rationale content preserves
a question-relevant distinction that is lost when reports are reduced to
verdict labels. The quantitative decomposition in
Section~\ref{sec:aggregation} separately measures the aggregate
contribution of joint comparison over verdicts and of restoring rationale
content.

\section{Design and Inference Details}
\label{app:design_inference}

This section gives the detailed comparisons underlying the design and
inference analysis in the main text: collapsing all JCA operations into one
completion, merging candidate checks within Step~2, comparing staged JCA with
CoT under the tested reasoning regimes, and measuring inference cost.

\subsection{OnePass versus Staged JCA}
\label{app:onepass}

OnePass JCA performs constraint extraction, candidate checking, and final
selection within a single completion. Table~\ref{tab:onepass_full} reports
paired comparisons against staged JCA on the two instruction-model backbones
for which OnePass was evaluated.

\begin{table*}[t]
\centering
\small
\setlength{\tabcolsep}{5pt}
\caption{OnePass JCA relative to staged JCA on the full evaluation sets and
hard slices. Values are accuracy differences in percentage points,
OnePass $-$ staged JCA, with 95\% paired-bootstrap confidence intervals.
All twelve comparisons significantly favor staged JCA.}
\label{tab:onepass_full}
\begin{tabular}{llcc}
\toprule
\textbf{Dataset} & \textbf{Backbone}
& \textbf{Full $\Delta$ [95\% CI]}
& \textbf{Hard $\Delta$ [95\% CI]} \\
\midrule
AmbigQA
& Qwen3-30B
& $-4.02$ [$-6.03$, $-2.01$]
& $-10.20$ [$-14.29$, $-6.46$] \\
& Llama4-Scout-17B
& $-8.29$ [$-10.93$, $-5.78$]
& $-16.67$ [$-21.77$, $-11.56$] \\
\midrule
AmbigQA-Extracted
& Qwen3-30B
& $-4.38$ [$-6.75$, $-2.19$]
& $-9.60$ [$-14.14$, $-5.05$] \\
& Llama4-Scout-17B
& $-5.84$ [$-8.58$, $-3.28$]
& $-10.10$ [$-15.66$, $-5.05$] \\
\midrule
SituatedQA
& Qwen3-30B
& $-8.31$ [$-11.87$, $-4.90$]
& $-17.65$ [$-23.53$, $-11.76$] \\
& Llama4-Scout-17B
& $-10.39$ [$-13.80$, $-7.12$]
& $-14.29$ [$-20.17$, $-8.40$] \\
\bottomrule
\end{tabular}
\end{table*}

Collapsing the three operations into one completion costs approximately
4--10 points on the full evaluation sets and 10--18 points on the hard slices
in these instruction-model comparisons. This result concerns collapsing the
stages entirely. It does not imply that Step~2 must use separate calls for
individual candidates.

\subsection{Merged versus Staged Candidate Checking}
\label{app:merged}

Merged-check JCA retains separate constraint-extraction, candidate-checking,
and aggregation stages but evaluates all candidates together within Step~2.
We test non-inferiority to staged JCA with a prespecified margin of 2.5
percentage points. Table~\ref{tab:merged_full} reports full-set comparisons.
Because the three datasets are disjoint, the primary non-inferiority tests pool
instances across datasets within each backbone but not across
backbones.

\begin{table*}[t]
\centering
\small
\setlength{\tabcolsep}{4pt}
\caption{Merged-check JCA versus staged JCA on the full evaluation sets.
$\Delta$ is merged $-$ staged accuracy in percentage points. NI is certified
when the lower bound of the 95\% confidence interval exceeds $-2.5$ points, and
``Superior'' denotes an interval entirely above zero. Per-dataset rows are
secondary comparisons, while the pooled within-backbone rows provide the
primary non-inferiority tests.}
\label{tab:merged_full}
\begin{tabular}{llrcc}
\toprule
\textbf{Dataset} & \textbf{Backbone} & \textbf{$n$}
& \textbf{$\Delta$ [95\% CI]} & \textbf{Result} \\
\midrule
AmbigQA
& Qwen3-30B & 796
& $-0.13$ [$-1.88$, $+1.63$] & NI \\
& Gemma3-27B & 796
& $-0.13$ [$-2.76$, $+2.51$] & Inconclusive \\
& Llama4-Scout-17B & 796
& $-0.13$ [$-2.01$, $+1.76$] & NI \\
\midrule
AmbigQA-Extracted
& Qwen3-30B & 548
& $-1.64$ [$-3.28$, $+0.00$] & Inconclusive \\
& Gemma3-27B & 548
& $-0.36$ [$-3.47$, $+2.74$] & Inconclusive \\
& Llama4-Scout-17B & 548
& $-1.28$ [$-3.28$, $+0.73$] & Inconclusive \\
\midrule
SituatedQA
& Qwen3-30B & 674
& $\mathbf{+3.71}$ [$+0.89$, $+6.53$] & Superior \\
& Gemma3-27B & 674
& $+0.59$ [$-3.12$, $+4.15$] & Inconclusive \\
& Llama4-Scout-17B & 674
& $+1.93$ [$-0.59$, $+4.45$] & NI \\
\midrule
\multicolumn{5}{l}{\textit{Pooled within backbone}} \\
Qwen3-30B & & 2{,}018
& $+0.74$ [$-0.50$, $+2.03$] & NI \\
Gemma3-27B & & 2{,}018
& $+0.05$ [$-1.73$, $+1.83$] & NI \\
Llama4-Scout-17B & & 2{,}018
& $+0.25$ [$-1.04$, $+1.54$] & NI \\
\bottomrule
\end{tabular}
\end{table*}

Merged-check JCA is non-inferior to staged JCA in all three pooled
within-backbone comparisons. At the individual dataset--backbone level, four
of nine comparisons certify non-inferiority, including one significantly
positive comparison, and none is significantly worse. The results therefore
distinguish merging candidate checks within Step~2 from collapsing all three
operations into OnePass.

\subsection{Interaction with Reasoning Mode}
\label{app:reasoning_mode}

We compare staged JCA with CoT under Qwen3-30B using its Instruct (reasoning
off) and Thinking (reasoning on) checkpoints. Table~\ref{tab:reasoning_staging} reports hard-slice
paired differences. The reasoning-off comparison runs CoT and staged JCA with the Instruct
checkpoint, while the reasoning-on comparison runs both with the Thinking
checkpoint.

\begin{table*}[t]
\centering
\small
\setlength{\tabcolsep}{6pt}
\caption{Staged JCA relative to CoT under the two tested Qwen3-30B reasoning
regimes on the hard slice. Differences are staged JCA $-$ CoT in percentage
points. NI refers to staged JCA relative to CoT, both with the Thinking checkpoint, at
a 2.5-point non-inferiority margin.}
\label{tab:reasoning_staging}
\begin{tabular}{lccc}
\toprule
\textbf{Dataset}
& \textbf{Reasoning off $\Delta$ [95\% CI]}
& \textbf{Reasoning on $\Delta$ [95\% CI]}
& \textbf{Reasoning-on NI} \\
\midrule
AmbigQA
& $\mathbf{+7.14}$ [$+3.74$, $+10.88$]
& $+1.02$ [$-2.04$, $+4.08$]
& Yes \\
AmbigQA-Extracted
& $\mathbf{+8.59}$ [$+4.55$, $+13.13$]
& $+1.01$ [$-4.55$, $+6.57$]
& Not certified \\
SituatedQA
& $\mathbf{+20.59}$ [$+14.29$, $+26.89$]
& $+2.94$ [$-0.84$, $+6.72$]
& Yes \\
\bottomrule
\end{tabular}
\end{table*}

With reasoning disabled, staged JCA exceeds CoT significantly on all three
hard slices. With reasoning enabled, the differences shrink to 1.0--2.9
points and all three confidence intervals include zero. With the Thinking
checkpoint, staged JCA is non-inferior to CoT on AmbigQA and SituatedQA at
the 2.5-point margin, while the AmbigQA-Extracted interval is too wide to certify
non-inferiority. These comparisons are specific to Qwen3-30B and do not
establish a general relationship between explicit JCA staging and native
reasoning.

\subsection{Cost Measurement Details}
\label{app:cost_measurement}

Table~\ref{tab:cost_measurements} reports the inference-cost measurements used
in the design and inference analysis. We ran a dedicated Qwen3-30B-A3B
inference pass over 25 AmbigQA and 25 SituatedQA instances, using the Instruct
checkpoint for JCA and NLI and the Thinking checkpoint for CoT-thinking,
using greedy decoding with one request in flight during measurement. Costs are
first aggregated within each instance and then averaged separately for each
dataset, without pooling AmbigQA and SituatedQA. No corresponding cost
measurement was collected for AmbigQA-Extracted.

Sequential latency is the sum of the wall-clock times of all measured calls
for an instance. For methods with independent per-candidate calls, we also
report candidate-parallel latency, estimated as the maximum latency among the
candidate-level calls plus the latency of stages that must remain sequential.
Staged JCA and NLI have such a parallelizable candidate-level group.
Merged-check JCA performs all candidate checks in one call, while
CoT-thinking uses a single call, so their candidate-parallel and sequential
latencies are identical by construction. Because candidate-parallel latency is
derived from sequentially measured call latencies rather than measured under
concurrent execution, it does not account for server-side contention or other
effects of actual concurrent serving.

The Step~1 constraint-extraction call was not instrumented in this cost pass.
Accordingly, the input- and output-token columns include the candidate-checking
and answer stages but not constraint extraction, whereas the request count
includes the Step~1 call. The omitted stage is shared by staged and
merged-check JCA, so it does not affect their relative input-token comparison.
Comparisons of JCA token totals with single-call methods therefore omit this
small JCA-only stage.

\begin{table*}[t]
\centering
\small
\begin{tabular}{llrrrrr}
\toprule
\textbf{Dataset} & \textbf{Method}
& \textbf{Req./inst.}
& \textbf{Input tok.}
& \textbf{Output tok.}
& \textbf{Seq. lat. (s)}
& \textbf{Cand.-par. lat. (s)} \\
\midrule
AmbigQA
& NLI            & 2.24 & 94,069 & 4     & 9.16  & 8.91 \\
& Staged JCA     & 4.24 & 96,030 & 733   & 28.77 & 13.85 \\
& Merged-check JCA & 3.00 & 43,090 & 888   & 19.32 & 19.32 \\
& CoT-thinking   & 1.00 & 41,485 & 1,424 & 23.85 & 23.85 \\
\midrule
SituatedQA
& NLI            & 2.00 & 82,274 & 4     & 9.14  & 8.94 \\
& Staged JCA     & 4.00 & 84,198 & 872   & 27.71 & 15.27 \\
& Merged-check JCA & 3.00 & 42,818 & 994   & 20.33 & 20.33 \\
& CoT-thinking   & 1.00 & 41,139 & 2,275 & 34.58 & 34.58 \\
\bottomrule
\end{tabular}
\caption{
Average per-instance inference cost on 25 AmbigQA and 25 SituatedQA
instances using Qwen3-30B-A3B with greedy decoding (Instruct checkpoint for JCA
and NLI, Thinking checkpoint for CoT-thinking).
Measurements were collected with one request in flight.
Candidate-parallel latency is derived from the measured call latencies as
described in Appendix~\ref{app:cost_measurement} and is not a wall-clock measurement under
concurrent serving. Step~1 constraint-extraction tokens were not instrumented,
although its request is included in the request count.
}
\label{tab:cost_measurements}
\end{table*}

These measurements illustrate different inference operating points rather
than a single uniformly preferable implementation. Relative to staged JCA,
merged-check JCA reduces measured input tokens by 55.1\% on AmbigQA and
49.1\% on SituatedQA, while also reducing the number of requests and
sequential latency. Staged JCA instead preserves independent candidate checks
that can be dispatched concurrently, yielding lower candidate-parallel latency
than merged-check JCA on both measured datasets. Latency values should be
interpreted as deployment-specific measurements from this inference setup
rather than as model-independent runtime estimates.

\section{Additional Baselines}
\label{app:baselines}

\subsection{Self-Consistency over CoT}
\label{app:self_consistency}

We evaluate whether additional sampling improves the CoT baseline on AmbigQA
with Qwen3-30B. For an instance with $C$ candidates, CoT-SC draws $K=C$
independent CoT samples at temperature $0.7$, each with a freshly shuffled
candidate order, and selects the candidate receiving
the majority vote, resolving voting ties by average answer log-probability.
The reported text-output implementation uses the serving default for top-$p$
and no fixed sampling seed.

On the final AmbigQA evaluation set, CoT, CoT-SC, and staged JCA achieve
90.7\%, 91.7\%, and 94.0\% accuracy on the full set, respectively, and
86.4\%, 86.7\%, and 93.5\% on the hard slice. Relative to single-sample CoT,
self-consistency changes accuracy by $+1.01$ points on the full set
(95\% CI $[-0.38,+2.39]$) and $+0.34$ points on the hard slice
($[-2.72,+3.06]$), and neither difference is statistically significant.
CoT-SC remains below staged JCA by 2.26 points on the full set
($[-4.15,-0.38]$) and 6.80 points on the hard slice
($[-10.88,-3.06]$). Additional sampling therefore does not recover staged
JCA's accuracy in this setting, although $K=C$ is a small sample budget (median 2).

\subsection{Cross-Encoder Rerankers}
\label{app:reranker}

We also evaluate two backbone-independent cross-encoder rerankers:
\nolinkurl{cross-encoder/ms-marco-MiniLM-L-6-v2}
\citep{wang2020minilm,bajaj2016msmarco} and
\nolinkurl{BAAI/bge-reranker-large} \citep{xiao2024cpack}. For each question--candidate pair, the
reranker scores each retrieved passage separately. We aggregate passage-level
scores for a candidate using either their maximum or their mean, then select
the candidate with the highest aggregate score. The rerankers are used
off-the-shelf without task-specific fine-tuning.
Table~\ref{tab:reranker_variants} reports both
aggregation variants.

\begin{table*}[t]
\centering
\small
\setlength{\tabcolsep}{5pt}
\caption{Cross-encoder reranker accuracy (\%) on the final evaluation sets.
``Max'' and ``Mean'' aggregate passage-level scores for each candidate by the
corresponding operation. Full/hard set sizes are 796/294 for AmbigQA,
548/198 for AmbigQA-Extracted, and 674/238 for SituatedQA.}
\label{tab:reranker_variants}
\begin{tabular}{llrrrr}
\toprule
\textbf{Reranker} & \textbf{Dataset}
& \textbf{Max Full} & \textbf{Max Hard}
& \textbf{Mean Full} & \textbf{Mean Hard} \\
\midrule
MiniLM
& AmbigQA            & 59.3 & 33.0 & 54.6 & 19.4 \\
& AmbigQA-Extracted  & 57.7 & 41.9 & 52.4 & 24.2 \\
& SituatedQA         & 55.0 & 27.3 & 52.5 & 25.2 \\
\midrule
BGE-large
& AmbigQA            & 59.5 & 41.5 & 50.6 & 20.4 \\
& AmbigQA-Extracted  & 56.4 & 52.3 & 49.8 & 27.3 \\
& SituatedQA         & 55.8 & 44.5 & 54.3 & 36.6 \\
\bottomrule
\end{tabular}
\end{table*}

Maximum aggregation outperforms mean aggregation for both rerankers in every
dataset and slice. BGE-large improves hard-slice accuracy over MiniLM by
8.5, 10.4, and 17.2 points on AmbigQA, AmbigQA-Extracted, and SituatedQA,
respectively. It nevertheless remains 51.0, 39.6, and 35.3 points below the
primary Llama4-Scout-17B staged-JCA results on those same hard slices. Its
full-set accuracy does not consistently improve over MiniLM. Stronger
cross-encoder reranking therefore narrows the reranking gap on the diagnostic
hard slice but does not approach JCA's selection accuracy in these
experiments.

\end{document}